\documentclass[10pt,twocolumn,letterpaper]{article}

\usepackage{cvpr}
\usepackage{times}
\usepackage{epsfig}
\usepackage{graphicx}
\usepackage{amsmath}
\usepackage{amssymb}

\usepackage{enumitem}
\usepackage{tablefootnote}
\usepackage{multirow}
\usepackage{capt-of}
\usepackage{fixltx2e}
\usepackage{floatrow}
\usepackage{pifont}

\usepackage[pagebackref=true,breaklinks=true,letterpaper=true,colorlinks,bookmarks=false]{hyperref}

\cvprfinalcopy 
\def\cvprPaperID{480} 

\ifcvprfinal\fi

\makeatletter
\newcommand{\printfnsymbol}[1]{%
	\textsuperscript{\@fnsymbol{#1}}%
}
\makeatother

\begin{document}

\title{Syn2Real Transfer Learning for Image Deraining using Gaussian Processes}

\author{Rajeev Yasarla\thanks{equal contribution} \qquad Vishwanath A. Sindagi\printfnsymbol{1} \qquad Vishal M. Patel\\
	Johns Hopkins University\\
	Department of Electrical and Computer Engineering, Baltimore, MD 21218, USA\\
	{\tt\small \{ryasarl1, vishwanathsindagi, vpatel36\}@jhu.edu}
}

\maketitle

\begin{abstract}
 Recent CNN-based methods for image deraining  have achieved excellent performance in terms of reconstruction error as well as visual quality. However, these methods are limited in the sense that they can be trained only on fully labeled data. Due to various challenges in obtaining real world fully-labeled image deraining datasets, existing methods are trained only on synthetically generated data and hence, generalize poorly to real-world images. The use of real-world data in training image deraining networks is relatively less explored in the literature.  We propose a Gaussian Process-based semi-supervised learning framework which enables the network in learning to derain using synthetic dataset while generalizing better using  unlabeled real-world images. Through extensive experiments and ablations on several challenging datasets (such as Rain800, Rain200H and DDN-SIRR), we show that the proposed method, when trained on limited labeled data, achieves on-par performance with fully-labeled training. Additionally, we demonstrate that using unlabeled real-world images in the proposed GP-based framework results in superior performance as compared to existing methods. Code is available at: \url{https://github.com/rajeevyasarla/Syn2Real}.
 

\end{abstract}

\section{Introduction}
Images captured under rainy conditions are often of poor quality. The artifacts introduced by rain streaks adversely affect the performance of subsequent computer vision algorithms such as object detection and recognition \cite{girshick2015fast,liu2016ssd, ren2015faster, Chen2018DomainAF}. With such algorithms becoming vital components in several applications such as autonomous navigation and video surveillance \cite{qi2018frustum,liang2018deep,perera2018uav}, it is increasingly important to develop sophisticated algorithms for rain removal.

\begin{figure}[t!]
	\begin{center}
		\includegraphics[width=.323\linewidth,height=0.242\linewidth]{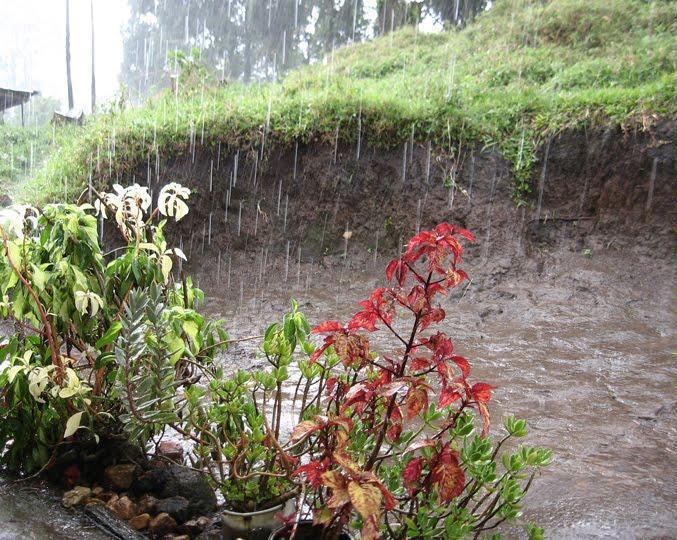}
		\includegraphics[width=.323\linewidth,height=0.242\linewidth]{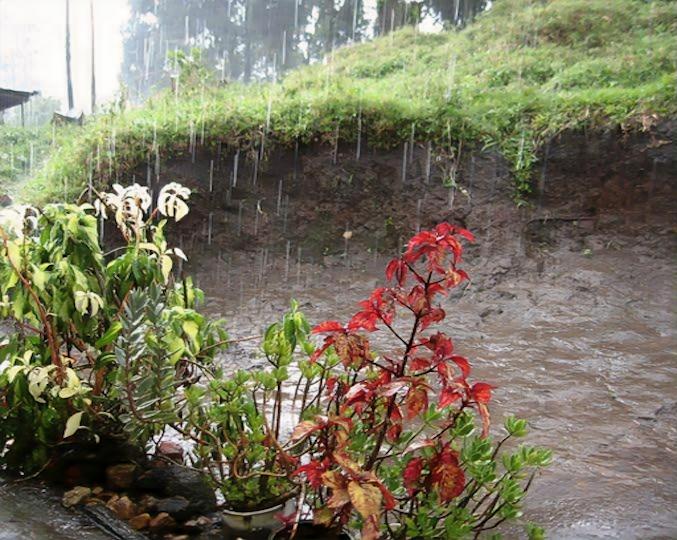}
		\includegraphics[width=.323\linewidth,height=0.242\linewidth]{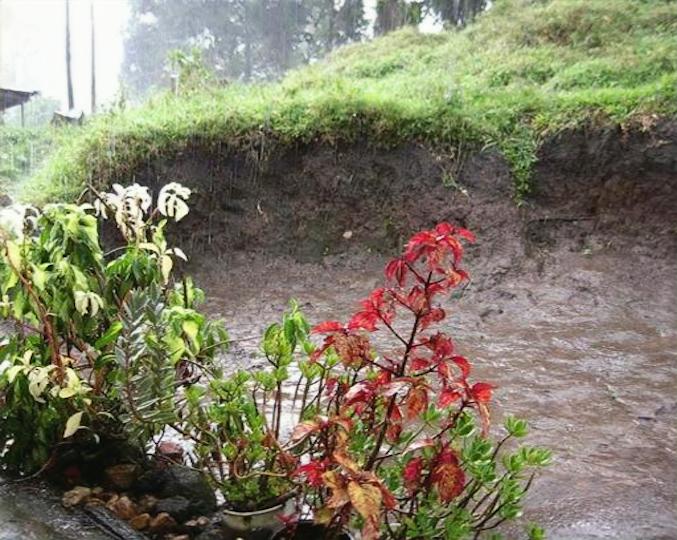}\\ 
	 	\includegraphics[width=.323\linewidth]{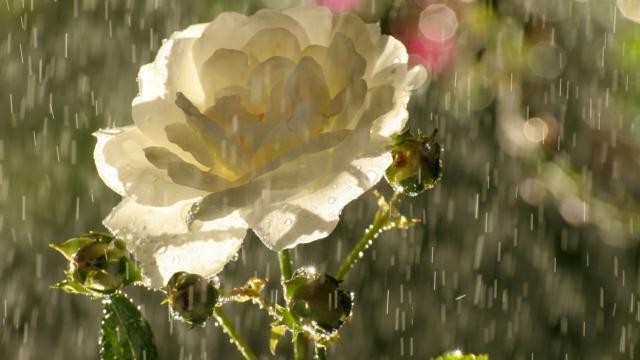}
	 	\includegraphics[width=.323\linewidth]{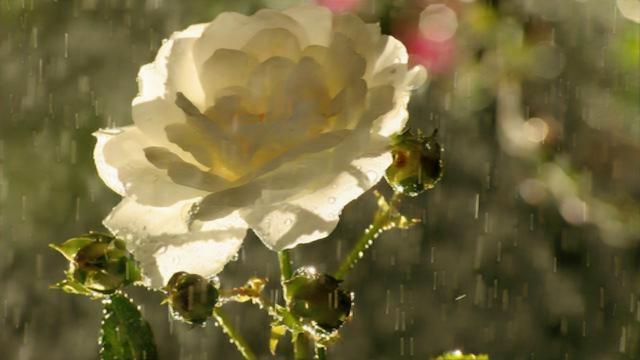}
	 	\includegraphics[width=.323\linewidth]{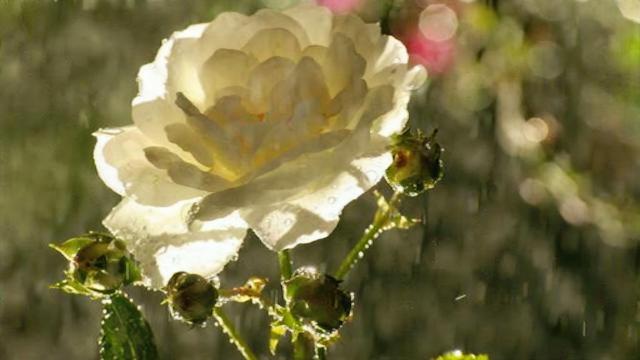}\\
		(a) \hskip60pt (b)  \hskip60pt (c)\\  
		\includegraphics[width=1\linewidth]{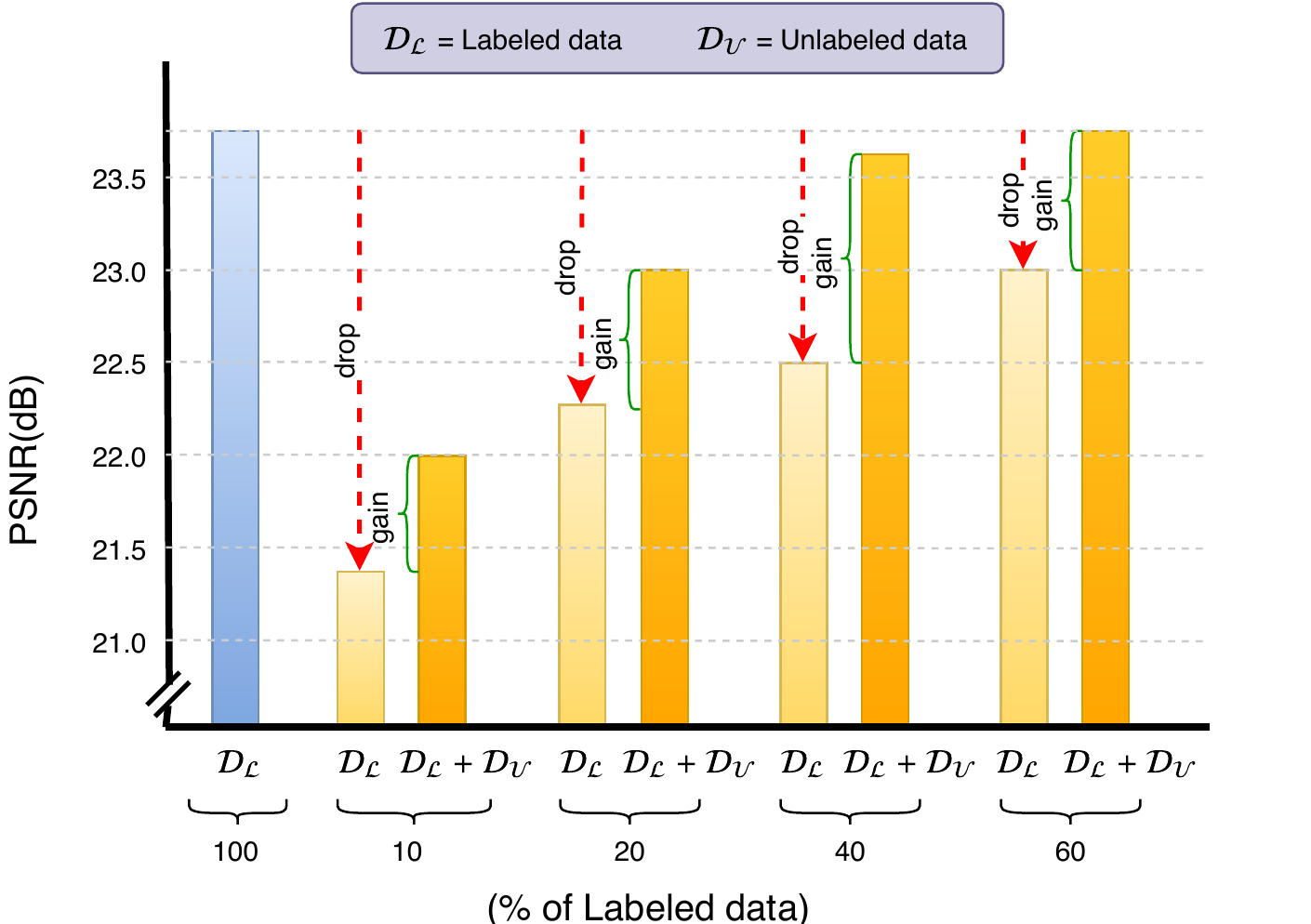}
	 
	\end{center}
	\vskip -15pt \caption{\textit{Top row}: (a) Input rainy image. (b) Output from a network trained using only the synthetic data. (c)  Output from a network trained using the synthetic data and unlabeled real-world data. This shows better generalization. \textit{Bottom row}: Results from Semi-supervised learning (SSL)  experiments. Reducing the amount of labeled data used for training results in the performance drop. Using the proposed SSL framework, we are able to recover the performance.}
	\label{fig:img1}
\end{figure}

The task of rain removal is plagued with several issues such as (i) large variations in scale, density and orientation of the rain streaks, and (ii) lack of real-world labeled training data. Most of the existing work \cite{Authors18,Authors17f,Authors16,yang2017deep, Authors17c,fan2018residual,yang2019joint,li2019single} in image deraining have largely focused towards addressing the first issue. For example, Fu \etal \cite{Authors17f} developed an end-to-end method which focuses on high frequency detail during training a deraining network.  In another work, Zhang and Patel \cite{Authors18} proposed a density-aware multi-steam densely connected network for joint rain density estimation and deraining. Li \etal \cite{Authors18d} incorporated context information through recurrent neural networks for rain removal. More recently, Ren \etal \cite{ren2019progressive} introduced a progressive ResNet that leverages dependencies of features across stages.  While these methods have achieved superior performance in obtaining high-quality derained images, they are inherently limited due to the fact that they are fully-supervised networks and they can only leverage fully-labeled training data. However, as mentioned earlier, obtaining labeled real-world training data is quite challenging and hence, existing methods typically train their networks only on synthetically generated rain datasets \cite{Authors17e,yang2017deep}. 

The use of synthetic datasets results in sub-optimal performance on the real-world images, typically because of the distributional-shift between synthetic and rainy images \cite{Chen2018DomainAF}.  Despite this gap in performance, this issue remains relatively unexplored in the literature. 

\begin{figure}[t!]
	\begin{center}
		\includegraphics[width=.323\linewidth,height=0.242\linewidth]{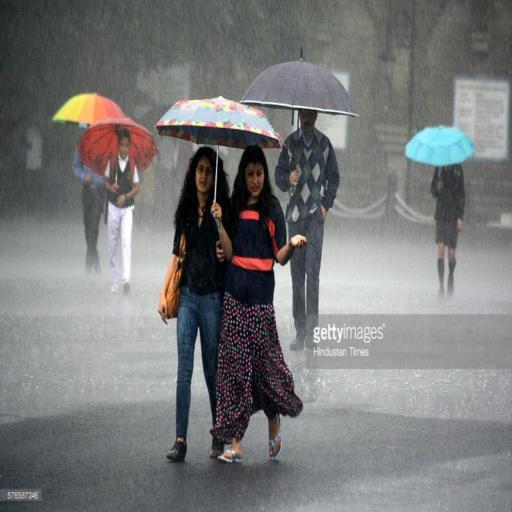}
		\includegraphics[width=.323\linewidth,height=0.242\linewidth]{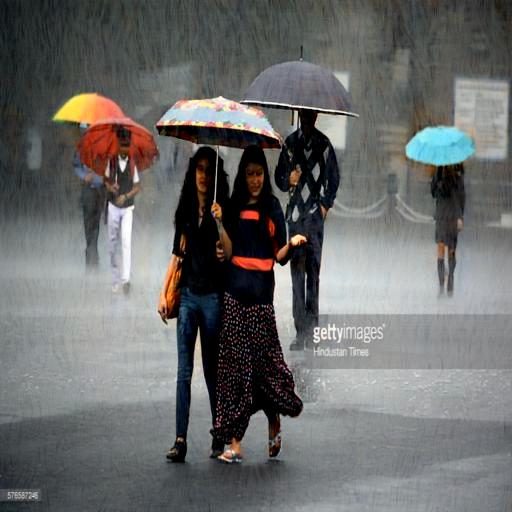}
		\includegraphics[width=.323\linewidth,height=0.242\linewidth]{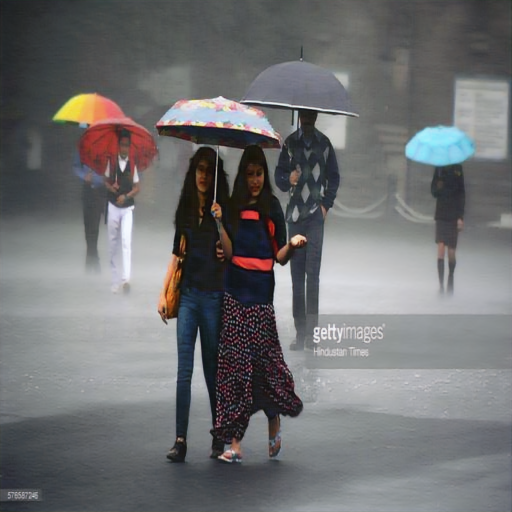}\\
		\includegraphics[width=.323\linewidth,height=0.242\linewidth]{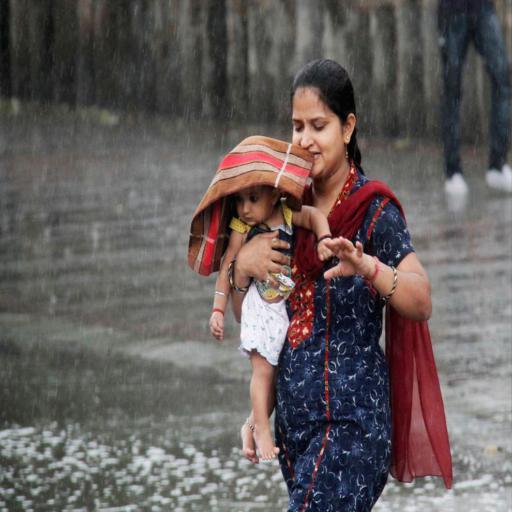}
		\includegraphics[width=.323\linewidth,height=0.242\linewidth]{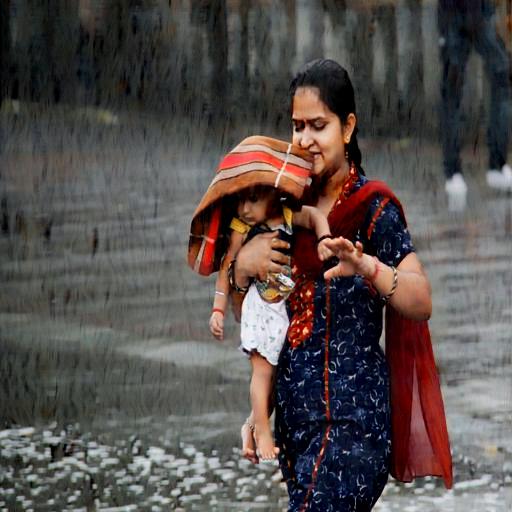}
		\includegraphics[width=.323\linewidth,height=0.242\linewidth]{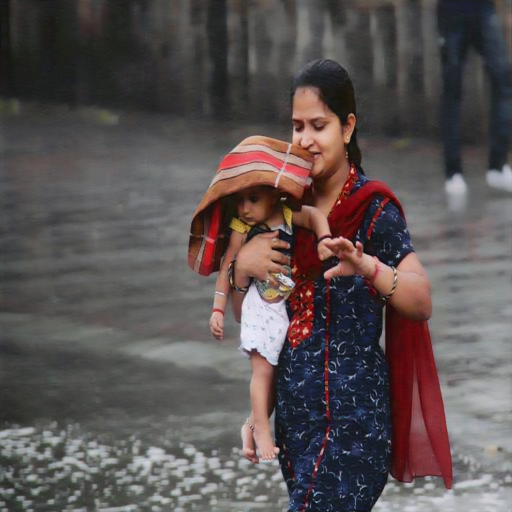}\\
		(a)\hskip 65pt(b)\hskip 65pt(c)
	\end{center}
	\caption{Derained results. (a) Input rainy images. (b)  SSIR output \cite{wei2019semi}. (c) Our output. It can be observed that the proposed method achieves better deraining.}
	\label{fig:img2}
\end{figure}

Recently, Wei \etal \cite{wei2019semi} proposed a semi-supervised learning framework (SIRR) where they simultaneously learn from labeled and unlabeled data for the purpose of image deraining. For training on the labeled data, they use the traditional mean absolute error loss between predictions and ground-truth (GT). For unlabeled data, they model the rain residual (difference between the input and output) through a likelihood term imposed on a Gaussian mixture model (GMM).  Furthermore, they enforce additional consistency that the distribution of synthetic rain is closer to that of real rain by minimizing the Kullback-Leibler (KL) divergence between them.  This is the first method to formulate the task of image deraining in a semi-supervised learning framework that can leverage unlabeled real-world images to improve the generalization capabilities. Although this method achieves promising results, it has the following drawbacks: (i) Due to the multi-modal nature of rain residuals, the authors assume that they  can be modeled using  GMM. This is true only if the actual residuals are being used to compute the GMM parameters. However, the authors use the  predicted rain residuals of real-world (unlabeled) images over training iterations for modeling the GMM. The  same model is then used to compute the likelihood of the predicted residuals (of unlabeled images) in the subsequent iterations. Hence, if the GMM parameters learned during the initial set of iterations are not accurate, which is most likely the case in the early stages of training, it will lead to sub-optimal performance. (ii) The goal  of using the KL divergence  is to bring the synthetic rain distribution  closer to the real rain distribution. As stated earlier, the predictions of real rain residuals will not be accurate during the earlier stages of training and hence, minimizing the discrepancy between the two distributions may not be appropriate. (iii) Using GMM to model the rain residuals requires one to choose the number of mixture components, rendering the model to be sensitive to such choices.

Inspired by Wei \etal \cite{wei2019semi}, we address the issue of incorporating unlabeled real-world images into the training process for better generalization by overcoming the drawbacks of their method. In contrast to \cite{wei2019semi}, we use a non-parametric approach to generate supervision for the unlabeled data.  Specifically, we propose a Gaussian-process (GP) based semi-supervised learning (SSL) framework which involves iteratively training on the labeled and unlabeled data. The labeled learning phase involves training on the labeled data using  mean squared error between the predictions and the ground-truth. Additionally, inputs (from labeled dataset)  are projected onto the latent space, which are then modeled using GP. During the unlabeled training phase, we generate pseudo-GT for the unlabeled inputs using the GP modeled earlier in the labeled training phase. This pseudo GT is then used to supervise the intermediate latent space for the unlabeled data. The creation of the pseudo GT is based on the assumption that unlabeled images, when projected to the latent space, can be expressed as a weighted combination of the labeled data features where the weights are determined using a kernel function. These weights indicate the uncertainty of the labeled data points being used to formulate the unlabeled data point. Hence,  minimizing the error between the unlabeled data projections and the pseudo GT reduces the variance, hence resulting in the network weights being adapted automatically to the domain of  unlabeled data. Fig. \ref{fig:img1} demonstrates the results of leveraging unlabeled data using the proposed framework. Fig. \ref{fig:img2} compares the results of the proposed method with SIRR \cite{wei2019semi}. One can clearly see that our method is able to provide better results as compared to SIRR \cite{wei2019semi}.

 To summarize, this paper makes the following contributions:
 \begin{itemize}[topsep=0pt,noitemsep,leftmargin=*]
 	\item We propose a non-parametric approach for performing SSL to  incorporate unlabeled real-world data into the training process. 
 	\item The proposed method consists of modeling the intermediate latent space in the network using GP, which is then used to create the pseudo GT for the unlabeled data. The pseudo GT is further used to supervise the network at the intermediate level for the unlabeled data. 
 	\item Through extensive experiments on different datasets,  we show that the proposed method is able to achieve on-par performance with limited training data as compared to network trained with full training data. Additionally, we also show that using the proposed GP-based SSL framework to incorporate the unlabeled real-world data into the training process results in better performance as compared to the existing methods. 
 \end{itemize}

\section{Related work}
Image deraining is an extensively researched topic in the low-level computer vision community. Several approaches have been developed  to address this problem. These approaches are  classified into two main  categories: single image-based techniques \cite{Authors18,Authors17f,Authors16,yang2017deep, Authors17c,9007569}  and video-based techniques \cite{Authors6,Authors7,Authors15c,Authors18e,Authors18f,liu2018d3r}.  A comprehensive analysis of these methods can be found in \cite{li2019singlecomprehensive}.

Single image-based techniques typically consume a single image as the input and attempt to reconstruct a rain-free image from it. Early methods for single image deraining either employed priors such as sparsity \cite{Authors17g, Authors15} and low-rank representation \cite{Low_Rank_ICCV2013} or modeled image patches using techniques such as dictionary learning \cite{Authors9} and GMM  \cite{Authors2000}. Recently, deep learning-based techniques have gained prominence due to their effectiveness in ability to learn efficiently from paired data. Video-based deraining techniques typically leverage additional information by enforcing  constraints like temporal consistency among the frames. 

In this work, we focus on single image-based deraining that specifically leverages additional unlabeled real-world data. Fu et al. \cite{Authors17d} proposed a convolutional neural network (CNN) based approach in which they learns a mapping  from a rainy image to  the clean  image.  Zhang et al. \cite{Authors17e} introduced  generative adversarial network (GAN) for image de-raining that resulted in high quality reconstructions.  Fu et al. \cite{Authors17f}  presented an end-to-end CNN called,    deep detail network, which directly reduces the mapping range from input to output.  Zhang and Patel \cite{Authors18}  proposed  a density-aware multi-stream densely connected CNN for joint rain density  estimation  and  deraining. Their network first classifies the input image based on the rain density, and then employs an appropriate network based on the predicted rain density to remove the rain streaks from the input image. Wang et al. \cite{Authors17b} employed  a hierarchical approach based on estimating different frequency details of an image to obtain the derained image.  Qian \etal \cite{Authors18b} proposed a GAN to remove rain drops from camera lens. To enable the network focus on important regions, they injected attention map into the generative and discriminating network.  Li et al. \cite{Authors18d} proposed a convolutional and recurrent neural network-based method for single image deraining that incorporates context information.    Recently, Li \etal \cite{li2019heavy}  and Hu \etal \cite{hu2019depth} incorporated depth information to improve the deraining quality. Yasarla and Patel \cite{yasarla2019uncertainty} employed uncertainty mechanism to learn location-based confidence for the predicted residuals. Wang \etal \cite{wang2019spatial} proposed a spatial attention network that removes rain in a local to global manner. 

 \section{Background}
 
 In this section, we provide a formulation of the problem statement, followed by a brief description of key concepts in GP.\\
 
 \subsection{Single image de-raining} 
 Existing image deraining methods assume the additive model where the rainy image ($x$) is considered to be the superposition of a clean image ($y$) and a rain component ($r$), \ie,
 \setlength{\belowdisplayskip}{0pt} \setlength{\belowdisplayshortskip}{0pt}
 \setlength{\abovedisplayskip}{0pt} \setlength{\abovedisplayshortskip}{0pt}
 \begin{equation}
 x= y + r.
 \end{equation}
Single image deraining task is typically an inverse problem where the goal is to   estimate the clean image $y$, given a rainy image $x$. This can be achieved by learning a function that either (i) directly maps from rainy image to clean image \cite{eigen2013restoring,fu2017clearing,Authors17c,Authors17g}, or (ii) extracts the rain component from the rainy image which can then be subtracted from the rainy image to obtain the clean image \cite{Authors17f,Authors18,li2018recurrent}. We follow the second approach of estimating the rain component from a rainy image.

\subsection{Semi-supervised learning}
 In  semi-supervised learning, we are given a labeled dataset of input-target pairs ($\{x,y\} \in \mathcal{D_L}$) sampled from an unknown joint distribution $p(x,y)$  and unlabeled input data points $x \in \mathcal{D_U}$ sampled from $p(x)$. The goal is to learn a  function $f(x|\theta)$  
parameterized by $\theta$ that accurately predicts the correct target $y$ for unseen samples from $p(x)$. The parameters $\theta$ are learned by leveraging both labeled and unlabeled datasets. Since the  labeled dataset  consists of input-target pairs, supervised loss functions such as mean absolute error or cross entropy are typically used to train the networks. The unlabeled datapoints form $\mathcal{D_U}$
 are used to augment $f(x|\theta)$   with information about the structure of $p(x)$ like shape of the data manifold \cite{oliver2018realistic} via different techniques such as enforcing consistent regularization \cite{laine2016temporal}, virtual adversarial training \cite{miyato2018virtual}  or pseudo-labeling \cite{lee2013pseudo}. 
 
Following \cite{wei2019semi}, we employ the semi-supervised learning framework to leverage unlabeled real-world data to obtain better generalization performance. Specifically, we consider the synthetically generated rain dataset consisting of input-target pairs as the labeled dataset $\mathcal{D_L}$ and real-world unlabeled images as the unlabeled dataset $\mathcal{D_U}$. In contrast to \cite{wei2019semi}, we follow the approach of pseudo-labeling to leverage the unlabeled data.

 \subsection{Gaussian processes}
 A Gaussian process (GP) $f(v)$ is an infinite collection of random variables, of which any finite subset is jointly Gaussian distributed. A GP is completely specified by its mean function and covariance function which are defined as follows
 \begin{equation}
 \begin{aligned} m(v) &=\mathbb{E}[f(v)],
 \end{aligned}
 \end{equation}
 \begin{equation}
   \begin{aligned} {K}\left(v, v^{\prime}\right) &=\mathbb{E}\left[(f(v)-m(v))\left(f\left(v^{\prime}\right)-m\left(v^{\prime}\right)\right)\right], \end{aligned}
 \end{equation}
where $v,v' \in \mathcal{V}$ denote the possible inputs that index the GP. The covariance matrix is constructed from a covariance function, or kernel, ${K}$ which expresses some prior notion of smoothness of the underlying function. GP can then be denoted as follows
 \begin{equation}
 	f(v) \sim \mathcal{GP}(m(v), K(v, v')+\sigma_{\epsilon}^2I).
 \end{equation}
where is I identity matrix and $\sigma_{\epsilon}^2$ is the variance of the additive noise.  Any collection of function values is then jointly Gaussian as follows
\begin{equation}
f(V)=\left[f\left(v_{1}\right), \ldots, f\left(v_{n}\right)\right]^{T} \sim \mathcal{N}\left(\mu, K(V, V')+\sigma_{\epsilon}^2I\right)
\end{equation}
with mean vector and covariance matrix defined by the GP as mentioned earlier. To make predictions at unlabeled points, one can compute a Gaussian posterior distribution in closed form by conditioning on the observed data. The reader is referred to \cite{rasmussen2003gaussian} for a  detailed review on GP.

 \section{Proposed method}
 
 
  \begin{figure}[t!]
 	\begin{center}
 		\includegraphics[width=1\linewidth]{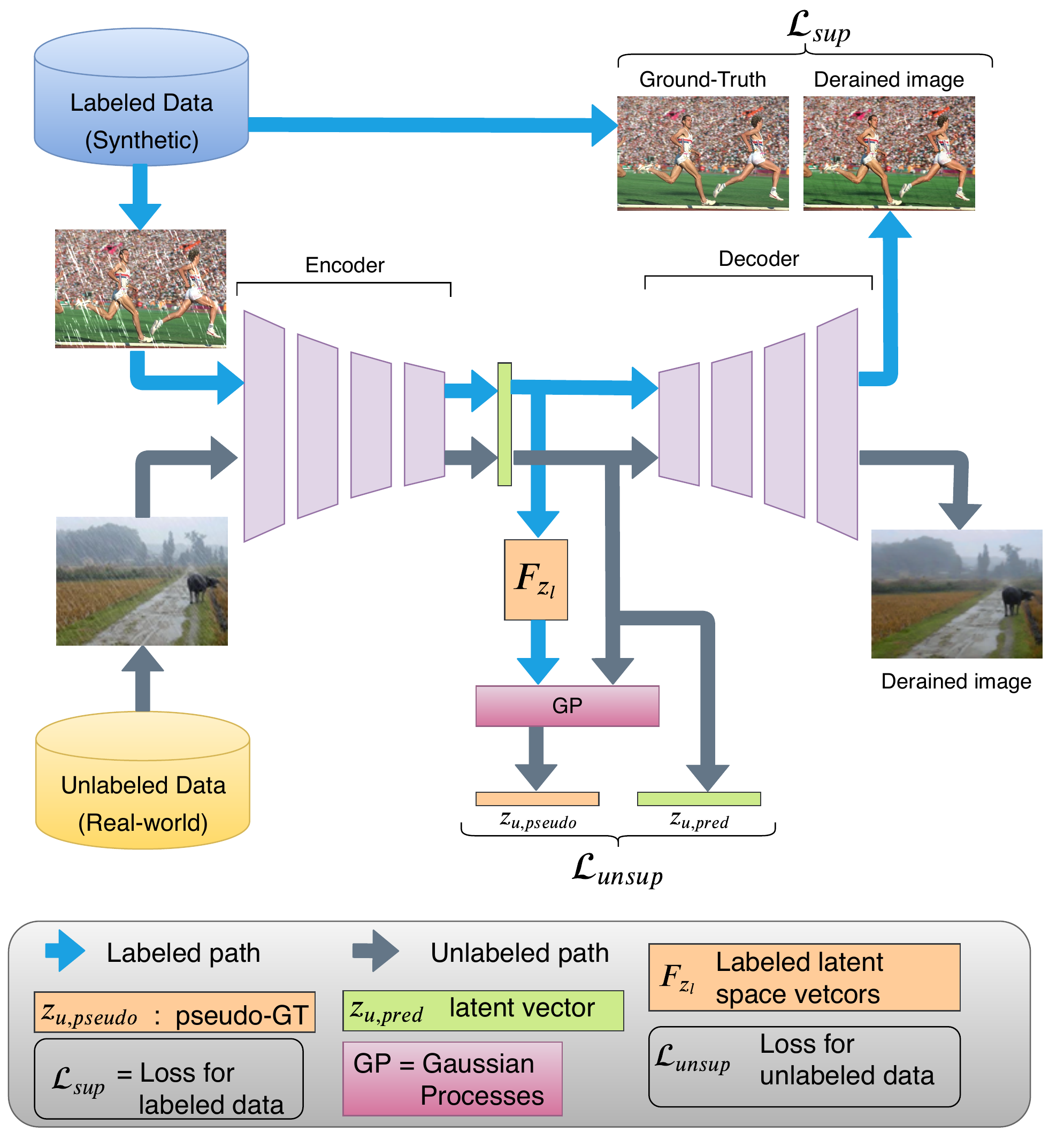} 
 	\end{center}
 	\vskip -10pt \caption{Overview of the proposed GP-based SSL framework. We leverage unlabeled data during learning. The training process consists of iterating over labeled data and unlabeled data. During the labeled training phase, we use supervised loss function consisting of $l_1$ error and perceptual loss between the prediction and targets. In the unlabeled phase, we jointly model the labeled and unlabeled latent vectors using GP to obtain the pseudo-GT for the unlabeled sample at the latent space. We use this pseudo-GT for supervision.}
 	\label{fig:overview}
 \end{figure}
 
 
As shown in Fig. \ref{fig:overview}, the proposed method consists of a CNN based on the UNet structure \cite{ronneberger2015u}, where each block is constructed using a Res2Block~\cite{gao2019res2net}. The details of the network architecture are provided in the supplementary material. In summary, the network is  made up of an encoder ($h(x,\theta_{enc})$) and a decoder ($g(z,\theta_{dec})$). Here, the encoder and decoder are parameterized by $\theta_{enc}$ and $\theta_{dec}$, respectively. Furthermore, $x$ is the input to the network  which is then mapped by the encoder to a latent vector $z$. In our case, $x$ is the rainy image from which we want to remove the rain streaks. The latent vector is then fed to the decoder to produce the output $r$, which in our case is the rain streaks. The rain streak component is then subtracted form the rainy image ($x$) to produce the clean image ($y$), \ie,
\begin{equation}
	y = x - r,
\end{equation}
where
\begin{equation} 
	r = g(h(x,\theta_{enc}), \theta_{dec}).
\end{equation}

 In our problem formulation, the training dataset is $\mathcal{D}=\mathcal{D_L} \cup \mathcal{D_U}$, where $\mathcal{D_L}=\{x_l^i,y_l^i\}_{i=1}^{N_l}$ is a labeled training set consisting of $N_{l}$ samples and $\mathcal{D_U}=\{x_u^i\}_{i=1}^{N_u}$ is a set consisting of $N_{u}$ unlabeled samples.  For the rest of the paper, $\mathcal{D_L}$ refers to   labeled ``\textit{synthetic}'' dataset and  $\mathcal{D_U}$ refers  to unlabeled ``\textit{real-world}'' dataset, unless otherwise specified. 
 
 The goal of the proposed method is to learn the network parameters by leveraging both labeled  ($\mathcal{D_L}$) and unlabeled dataset ($\mathcal{D_U}$). The training process iterates over labeled and unlabeled datasets. The network parameters are learned by minimizing (i) the supervised loss function ($\mathcal{L}_{sup}$) in the labeled training phase, and (ii) the unsupervised loss function ($\mathcal{L}_{unsup}$) in the unlabeled training phase.  For the unlabeled training phase, we generate pseudo GT using GP formulation, which is then used in the unsupervised loss function. The two training phases are described in detail in the following sections. 
 
\subsection{Labeled training phase}
In this phase, we use the labeled data $\mathcal{D_L}$ to  learn the network parameters. Specifically, we  minimize the following supervised loss function 
\begin{equation}
\label{eq:loss_sup}
\mathcal{L}_{sup} = \mathcal{L}_1 + \lambda_p \mathcal{L}_{p},
\end{equation}
where $\lambda_{p}$ is a constant, and $\mathcal{L}_1$ and $\mathcal{L}_p$ are $l_1$-loss and perceptual loss~\cite{johnson2016perceptual,zhang2018multi} functions, respectively. They are defined as follows
\begin{equation}
\mathcal{L}_{1} = \|y^{pred}_l - y_l\|_1,
\end{equation}
\begin{equation}
\mathcal{L}_{p} = \|\Phi_{VGG}(y^{pred}_l) - \Phi_{VGG}(y_l)\|^2_2 ,
\end{equation}
where $y^{pred}_l=g(z,\theta_{dec})$ is the predicted output,  $y_l$ is the ground-truth,  $z=h(x, \theta_{enc})$ is the intermediate latent space vector and $\Phi_{VGG}(\cdot)$ represents the pre-trained VGG-16 \cite{simonyan2014very} network. For more details on the perceptual loss, please refer to supplementary material. 

In addition to minimizing the loss function, we also store the intermediate feature vectors $z_l^i$'s for all the labeled training images $x_l^i$'s in a matrix $F_{z_l}$. That is $F_{z_l}= \{z_l^i\}_{i=1}^{N_l}$. It is used later in the unlabeled training phase to generate the pseudo-GT for the unlabeled data. In our case,  $z_l^i$ is a vector of size $1\times M$, where M = 32,768 for the network in our proposed method. Thus $F_{z_l}$ is a matrix of size  $N_l\times M$.
 
\subsection{Unlabeled training phase}
In this phase, we leverage the unlabeled data $\mathcal{D_U}$ to improve the generalization performance. Specifically, we provide supervision at the intermediate latent space by minimizing the error between the predicted latent vectors and the pseudo-GT  obtained by modeling the latent space vectors of the labeled sample images $F_{z_l}$ and $z^{pred}_u$ jointly using GP. \\

\noindent\textbf{Pseudo-GT using GP:} The training occurs in an iterative manner, where we first learn the weights using   the labeled data ($\mathcal{D_L}$) followed by weight updates using the unlabeled data ($\mathcal{D_U}$). After the first iteration on  $\mathcal{D_L}$, we store the latent space vectors of the labeled data  in a list $F_{z_l}$. These vectors lie on a low dimension manifold. During the unlabeled phase, we project the latent space vector ($z_u$) of the unlabeled input onto the space of labeled vectors $F_{z_l}= \{z_l^i\}_{i=1}^{N_l}$. That is, we express the unlabeled latent space vector $z^k_u$  corresponding to the $k^{th}$ training sample from $\mathcal{D_U}$ as
\begin{equation}
\label{eq:lin_combination}
z^k_u = \sum_{i=1}^{N_l} \alpha_i z_l^i +\epsilon,
\end{equation}
where $\alpha_{i}$ are the coefficients, and $\epsilon$ is additive noise $\mathcal{N}(0,\sigma_\epsilon^2)$.  

With this formulation, we can jointly model the distribution of the latent space vectors of the labeled and the unlabeled samples using GP. Conditioning the joint distribution will yield the following conditional multi-variate Gaussian distribution for the unlabeled sample
\begin{equation}
P(z_u^k|\mathcal{D_L},F_{z_l}) = \mathcal{N}(\mu_u^k,\Sigma_u^k),
\end{equation}
where 
\begin{equation}
\label{eq:mean}
\mu_u^k = K(z_u^k, F_{z_l}) [K(F_{z_l},F_{z_l}) + \sigma_\epsilon^2 I]^{-1}F_{z_l},
\end{equation}
\vskip2pt
\begin{equation}
\label{eq:sigma}
\begin{aligned}
\Sigma_u^k = {} & K(z_u^k,z_u^k) - K(z_u^k,F_{z_l})[K(F_{z_l},F_{z_l})+\sigma_\epsilon^2I]^{-1} \\
 & K(F_{z_l},z_u^k) + \sigma_\epsilon^2
\end{aligned}
\end{equation}
where $\sigma_\epsilon^2$ is set equal to 1, $K$  is defined by the kernel function as follows
\begin{equation}
K(Z,Z)_{k,i}= \kappa(z_u^k,z_l^i) = \frac{ \langle z_u^k, z_l^i\rangle}{|z_u^k|\cdot|z_l^i|}.
\end{equation}

Note that $F_{z_l}$ contains the latent space vectors of all the labeled images, $K(F_{z_l},F_{z_l})$ is a matrix of size $N_l\times N_l$, and $K(z_u^k,F_{z_l})$ is a vector of size $1\times N_l$. Using all the vectors may not be necessarily optimal for the following reasons: (i) These vectors will correspond to different regions in the image with a wide diversity in terms of content and density/orientation of rain streaks. It is important to consider only those vectors that are similar to the unlabeled vector.  (ii) Using all the vectors is computationally prohibitive. Hence, we use only $N_n$ nearest labeled vectors corresponding to an unlabeled vector. More specifically, we replace  $F_{z_l}$ by $F_{z_l,n}$ in Eq. \eqref{eq:lin_combination}-\eqref{eq:sigma}. Here $F_{z_l,n}=\{z_l^j :  z_l^j \in nearest(z_u^k,F_{z_l} ,N_n) \}$ with $nearest(p,Q ,N_n)$ being a function that finds top $N_n$ nearest neighbors of $p$ in $Q$.

\begin{table*}[ht!]
	\caption{Effect of using unlabeled real-world data in training process on DDN-SIRR dataset. Evaluation is performed on synthetic dataset similar to \cite{wei2019semi}. Proposed method achieves better gain in PSNR as compared to SIRR\cite{wei2019semi} in the case of both Dense and Sparse categories. $\mathcal{D_L}$ indicates training using only labeled dataset and $\mathcal{D_L + D_U}$ indicates training using both labeled and unlabeled dataset.}
	\vskip-8pt 
	\label{tab:ddnsirr_synthetic}
	\centering
	\resizebox{1\linewidth}{!}{
		\begin{tabular}{|l|c|ccccccc|ccc|ccc|}
			\hline
			\multirow{3}{*}{{Dataset}} & \multirow{3}{*}{\begin{tabular}[c]{@{}c@{}}Input\end{tabular}} & \multicolumn{7}{c|}{Methods that use only synthetic dataset}                                                                                  & \multicolumn{6}{c|}{Methods that use synthetic and real-world dataset} \\ \cline{3-15} 
			&                                                                         & \multirow{2}{*}{\begin{tabular}[c]{@{}c@{}}DSC \cite{luo2015removing}\\(ICCV '15)\end{tabular}} & \multirow{2}{*}{\begin{tabular}[c]{@{}c@{}}LP \cite{li2016rain}\\(CVPR '16)\end{tabular}} & \multirow{2}{*}{\begin{tabular}[c]{@{}c@{}}JORDER \cite{yang2017deep}\\(CVPR '17)\end{tabular}} & \multirow{2}{*}{\begin{tabular}[c]{@{}c@{}}DDN \cite{Authors17f}\\(CVPR '17)\end{tabular}} & \multirow{2}{*}{\begin{tabular}[c]{@{}c@{}}JBO \cite{Authors17c}\\(CVPR '17)\end{tabular}} & \multirow{2}{*}{\begin{tabular}[c]{@{}c@{}}DID-MDN \cite{Authors18}\\(CVPR '18)\end{tabular}} & \multirow{2}{*}{\begin{tabular}[c]{@{}c@{}}UMRL \cite{yasarla2019uncertainty}\\(CVPR '19)\end{tabular}} & \multicolumn{3}{c|}{SIRR \cite{wei2019semi} (CVPR '19)}  & \multicolumn{3}{c|}{Ours}                 \\ \cline{10-15} 
			&                                                                         &                      &                     &                         &                      &                      &                          & & $\mathcal{D_L}$     & $\mathcal{D_L+D_U}$  & Gain  & $\mathcal{D_L}$    & $\mathcal{D_L+D_U}$        & Gain           \\ \hline
			Dense                    & 17.95                                                                   & 19.00                & 19.27               & 18.75                   & 19.90                & 18.87                & 18.60   & 20.11                 & 20.01   & 21.60    & 1.59  & 20.24  & \textbf{22.36}  & \textbf{2.12}  \\
			Sparse                   & 24.14                                                                   & 25.05                & 25.67               & 24.22                   & 26.88                & 25.24                & 25.66  &  26.94      & 26.90   & 26.98    & 0.08  & 26.15  & \textbf{27.26}  & \textbf{1.11}  \\ \hline
		\end{tabular}
	}
	\vskip-5pt
\end{table*}
 
We use the mean predicted by Eq. \eqref{eq:mean} as the pseudo-GT ( $z_{u,pseudo}^{k}$) for supervision at the latent space level. By minimizing the error between $z^{k}_{u,pred}=h(x_u,\theta_{enc})$ and  $z_{u,pseudo}^{k}$, we update the weights of the encoder $h(\cdot,\theta_{enc})$, thereby adapting the network to unlabeled data which results in better generalization. We also minimize the prediction variance by  minimizing Eq. \eqref{eq:sigma}. Using GP we are approximating $z_{u}^{k}$, latent vector of an unlabeled image using the latent space vectors in $F_{z_l}$, by doing this we may end up computing incorrect pseudo-GT predictions because of the dissimilarity between the latent vectors. This dissimilarity is due to different compositions in rain streaks like different densities, shapes, and directions of rain streaks. In order to address this issue we minimize the variance $\Sigma_{u,n}^k$ computed between $z^{k}_{u}$ and the $N_n$ nearest neighbors in the latent space vectors using GP. Additionally, we maximize the variance $\Sigma_{u,f}^{k}$ computed between $z^{k}_{u}$ and the  $N_f$ farthest vectors in the latent space using GP, in order to ensure that the latent vectors in $F_{z_l}$  are dissimilar to the unlabeled vector $z_u^k$ and do not affect the GP prediction, as defined below
\vskip1pt
\begin{equation}
\label{eq:sigma_far}
\begin{aligned}
\Sigma_{u,f}^{k} = {} & K(z_u^k,z_u^k) - K(z_u^k,F_{z_l,f})[K(F_{z_l,f},F_{z_l,f})+\sigma_\epsilon^2I]^{-1}\\
& K(F_{z_l,f},z_u^k) + \sigma_\epsilon^2,
\end{aligned}
\end{equation}
where $F_{z_l,f}$ is the matrix of $N_f$ labeled vectors that are farthest from $z_u^k$. 

Thus, the loss used during training using the unlabeled data is defined as follows
\begin{equation}
\mathcal{L}_{unsup} = \|{z}^{k}_{u,pred} - {z}_{u,pseudo}^{k}\|_2 + \log \Sigma_{u,n}^{k} + \log(1-\Sigma_{u,f}^{k}),
\end{equation}
where $z^{k}_{u,pred}$ is the latent vector obtained by forwarding an unlabeled input image $x_u^k$ through the encoder $h$, \ie, $z^{k}_{u,pred}=h(x_u,\theta_{enc})$ , $z_{u,pseudo}^{k} = \mu_u^k$ is the  pseudo-GT latent space vector (see Eq. \eqref{eq:mean}), and $\Sigma_{u,n}^{k}$ is the variance obtained by replacing $F_{z_l}$ in Eq. \eqref{eq:sigma} with $F_{z_l,n}$.
 
\subsection{Total loss}
The overall loss function used for training the network is defined as follows
\begin{equation}
\label{eq:loss_total}
 \mathcal{L}_{total} = \mathcal{L}_{sup} + \lambda_{unsup} \mathcal{L}_{unsup},
\end{equation} 
 where $\lambda_{unsup}$ is a pre-defined weight that controls the contribution from $\mathcal{L}_{sup}$ and $\mathcal{L}_{unsup}$.

\subsection{Training and implementation details}
We use the UDeNet network that is  based on the UNet style encoder-decoder architecture \cite{ronneberger2015u} with a slight difference in the building blocks. Details of the network architecture are provided in the supplementary material. The network  is trained using the Adam optimizer with a learning rate of 0.0002 and batchsize of 4 for a total of 60 epochs. Furthermore, we reduce the  learning rate by a factor of 0.5 at every 25 epochs. We use $\lambda_p=0.04$ (Eq. \eqref{eq:loss_sup}), $\lambda_{unsup}=1.5 \times 10^{-4}$ (Eq. \eqref{eq:loss_total}), $N_n=64$ and $N_f=64$. During training, the images are randomly cropped to the size of 256$\times$256. Ablation studies with different hyper-parameter values are provided in supplementary material.

 \section{Experiments and results}
 In this section, we present the details of the datasets and various experiments conducted to demonstrate the effectiveness of the proposed framework.  Specifically, we conducted two sets of experiments. In the first set, we analyze the effectiveness of using the unlabeled real-world data during training using the proposed framework. Here, we compare the performance of our method with a recent SSL framework for image deraining (SIRR) \cite{wei2019semi}. In the second set of experiments, we evaluate the proposed method by training it on different percentages of the labeled data. 
 
 \subsection{Datasets}
 \noindent\textbf{Rain800:} This dataset was introduced by Zhang \etal \cite{Authors17e} and it contains a total of 800 images. The train split consists of 700 real-world clean images, with 500 images chosen randomly from the first half of the UCID dataset \cite{schaefer2003ucid} and 200 images chosen randomly from the BSD-500 train set \cite{arbelaez2010contour}. The test set consists of a total of 100 images, with 50 images chosen randomly from the second half of the UCID dataset and the rest 50 chosen randomly from the test set of the BSD-500 dataset. The authors generate the corresponding rainy images by synthesizing rain-streaks of different intensities and orientations. \\
 
 \noindent\textbf{Rain200H:} Yang et al. \cite{yang2017deep} collected images from BSD200 \cite{martin2001database} to create 3 datasets:  Rain12, Rain200L and Rain200H. Following \cite{li2018recurrent}, we use the most difficult one, Rain200H, to evaluate our model. The images for the training set are collected from the BSD300 dataset. Rain streaks with different orientations are synthesized  using photo-realistic techniques. 	There are 1,800 synthetic image pairs in the Rain200H train set, and 200 pairs in the test set.\\
 
 \noindent\textbf{DDN-SIRR dataset:} Wei \etal \cite{wei2019semi} constructed a dataset consisting of labeled synthetic training set and unlabeled real-world dataset. This dataset is constructed specifically to evaluate semi-supervised learning frameworks. The labeled training set is borrowed from Fu \etal \cite{Authors17f} and it  consists of 9,100 image pairs obtained by synthesizing  different types of rain streaks on the clean images from the UCID dataset \cite{schaefer2003ucid}. The unlabeled real-world synthetic train set comprises of images collected from \cite{wei2017should,yang2017deep,Authors17e} and Google image search. Furthermore, the test set consists of two categories: (i) Dense rain streaks, and (ii) Sparse rain streaks Each test set  consists of 10 images. 
 
\subsection{Use of real-world data}
The goal of this experiment is to analyze the effect of using unlabeled real-world data along with labeled synthetic dataset in the training framework. Following the protocol set by \cite{wei2019semi}, we use the ``labeled synthetic" train set from the DDN-SIRR dataset as  $\mathcal{D_L}$ and the ``real-world" train set from the DDN-SIRR dataset as $\mathcal{D_U}$. Evaluation is performed on (i) Synthetic test set from DDN-SIRR,  and (ii) Real-world test set from DDN-SIRR.\\  

\begin{figure*}[t!]
	\begin{center}
		\includegraphics[width=.16\linewidth]{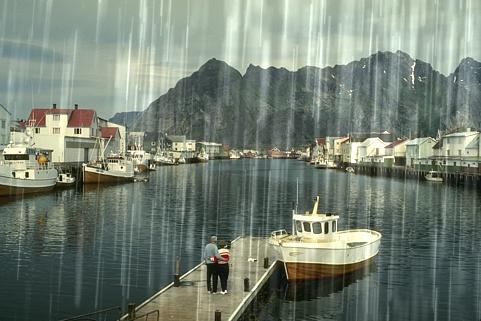}
		\includegraphics[width=.16\linewidth]{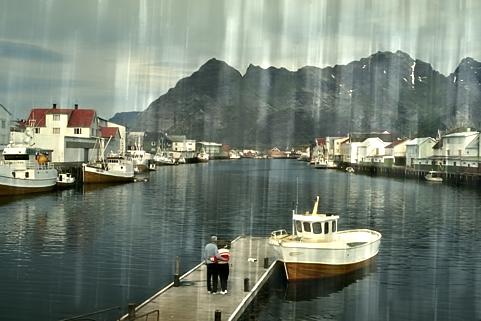}
		\includegraphics[width=.16\linewidth]{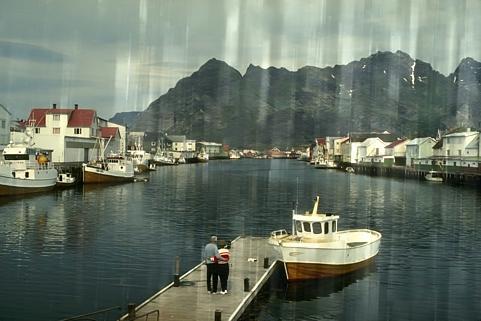}
		\includegraphics[width=.16\linewidth]{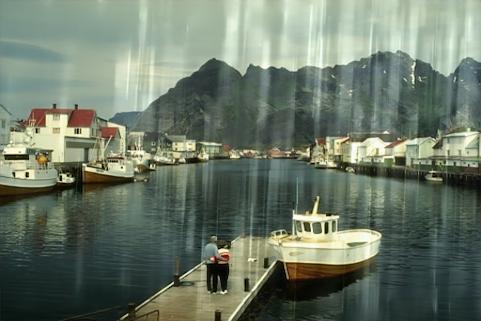}
		\includegraphics[width=.16\linewidth]{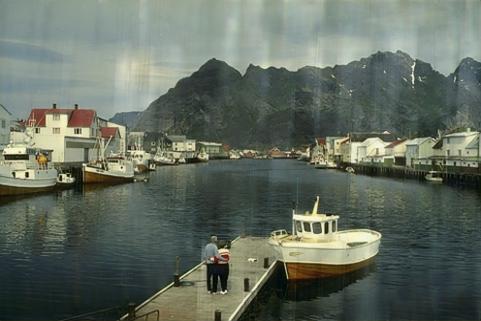}
		\includegraphics[width=.16\linewidth]{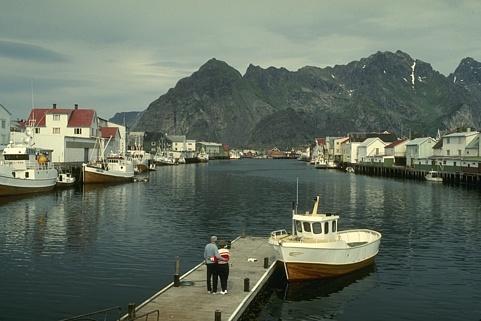}\\ 
		\includegraphics[width=.16\linewidth]{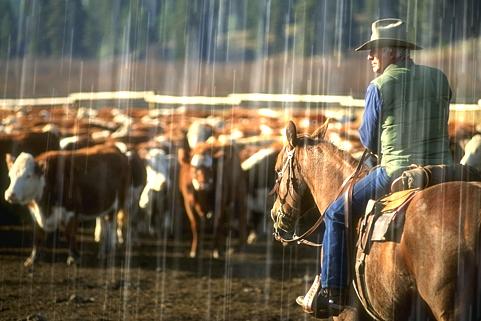}
		\includegraphics[width=.16\linewidth]{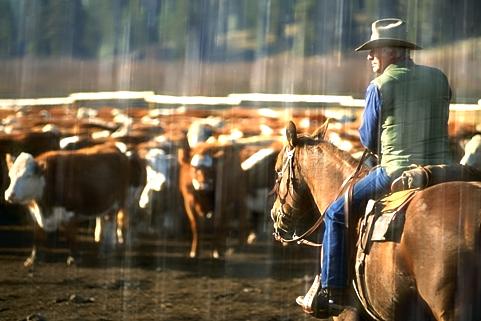}
		\includegraphics[width=.16\linewidth]{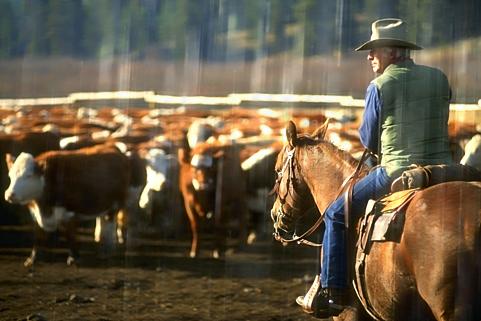}
		\includegraphics[width=.16\linewidth]{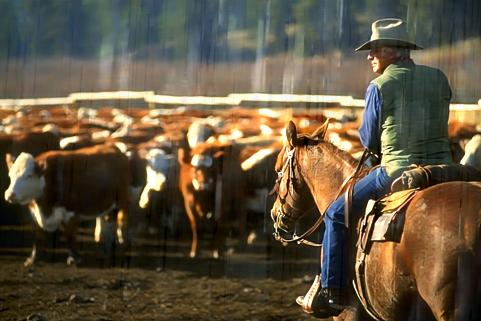}
		\includegraphics[width=.16\linewidth]{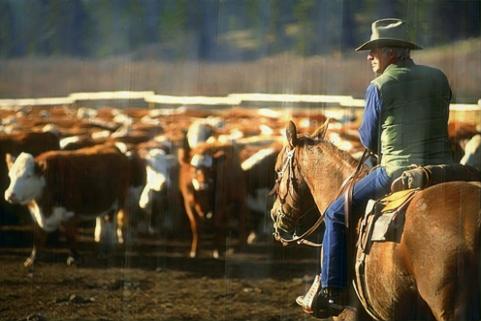}
		\includegraphics[width=.16\linewidth]{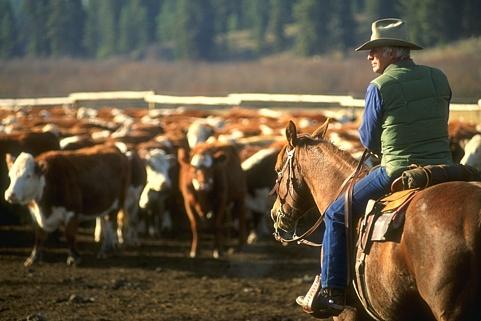}\\
		(a)\hskip 70pt(b)\hskip 70pt(c)\hskip 75pt(d)\hskip 70pt(e) \hskip 70pt (f)
	\end{center}
	\vskip -18pt \caption{Qualitative results on DDN-SIRR  \textbf{synthetic} test set. (a) Input rainy image (b) DID-MDN \cite{Authors18}(CVPR '18) (c) DDN \cite{Authors17f}(CVPR '17) (d) SIRR \cite{wei2019semi}(CVPR '19) (e) Ours (f) ground-truth image.}
	\label{fig:ddnsirr_synthetic}
\end{figure*}

\begin{figure*}[t!]
	\begin{center}
		\includegraphics[width=.195\linewidth]{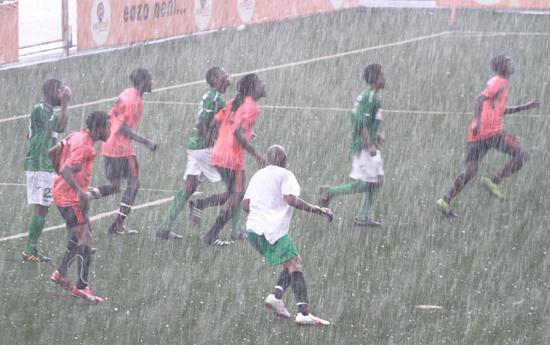}
		\includegraphics[width=.195\linewidth]{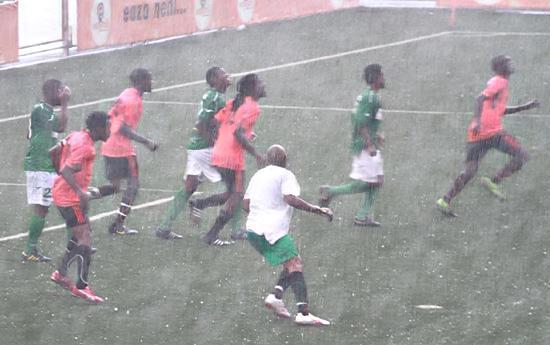}
		\includegraphics[width=.195\linewidth]{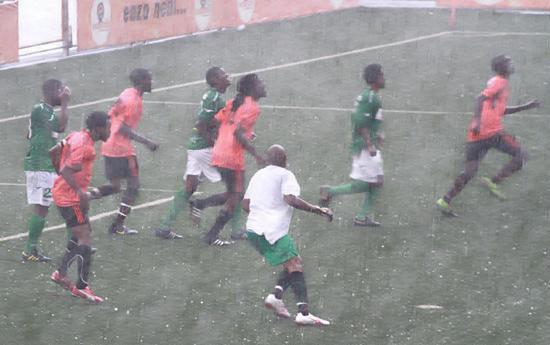}
		\includegraphics[width=.195\linewidth]{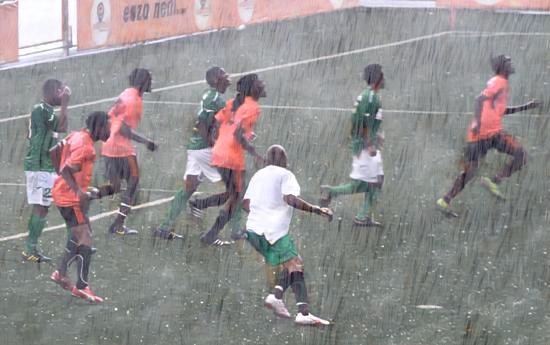}
		\includegraphics[width=.195\linewidth]{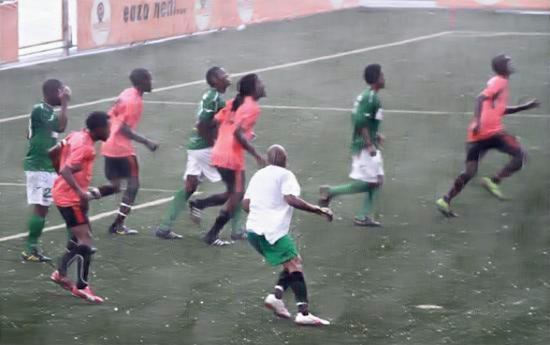}\\ 
		\includegraphics[width=.195\linewidth]{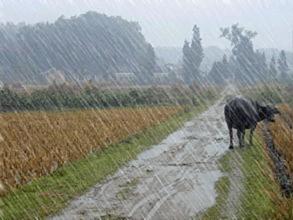}
		\includegraphics[width=.195\linewidth]{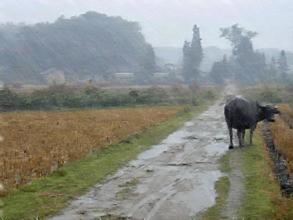}
		\includegraphics[width=.195\linewidth]{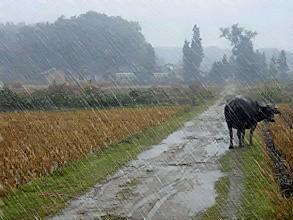}
		\includegraphics[width=.195\linewidth]{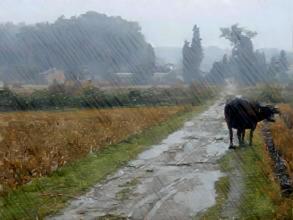}
		\includegraphics[width=.195\linewidth]{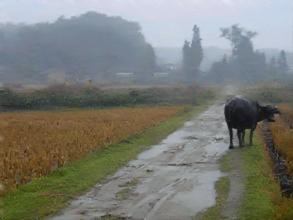}\\ 
		\includegraphics[width=.195\linewidth,height=0.146\linewidth]{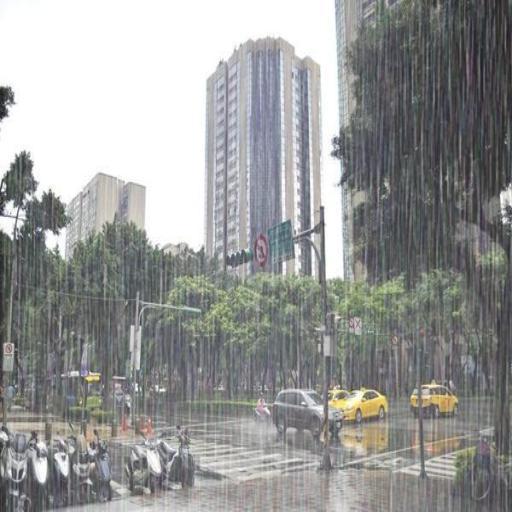}
		\includegraphics[width=.195\linewidth,height=0.146\linewidth]{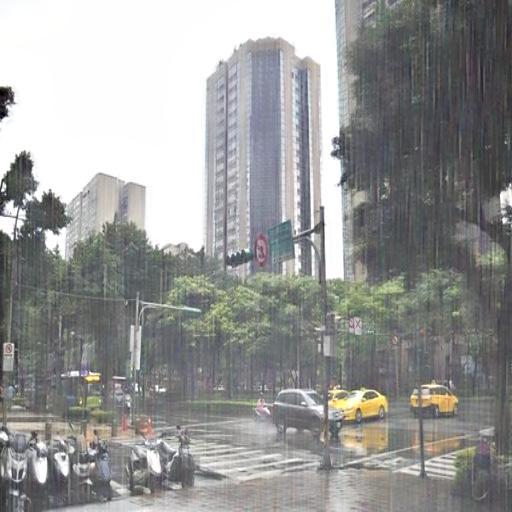}
		\includegraphics[width=.195\linewidth,height=0.146\linewidth]{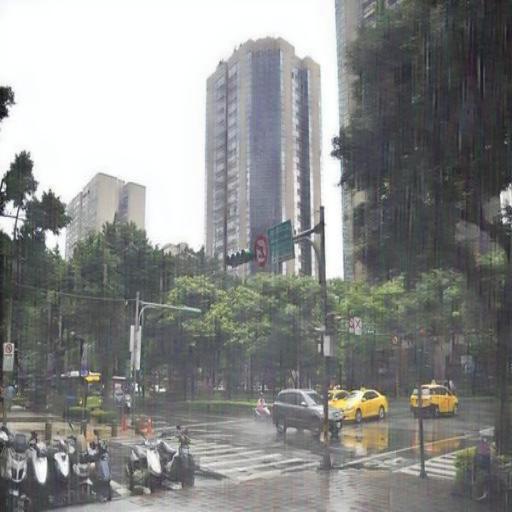}
		\includegraphics[width=.195\linewidth,height=0.146\linewidth]{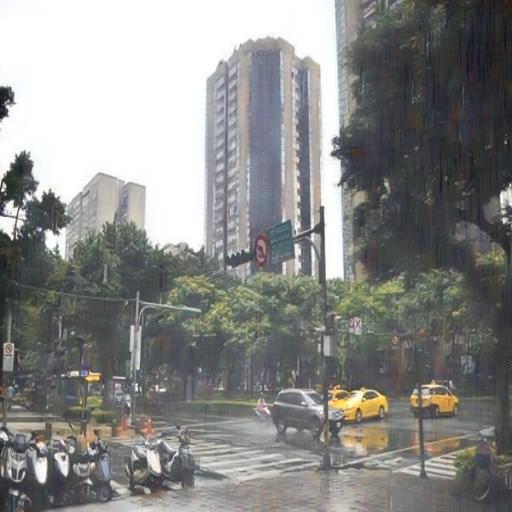}
		\includegraphics[width=.195\linewidth,height=0.146\linewidth]{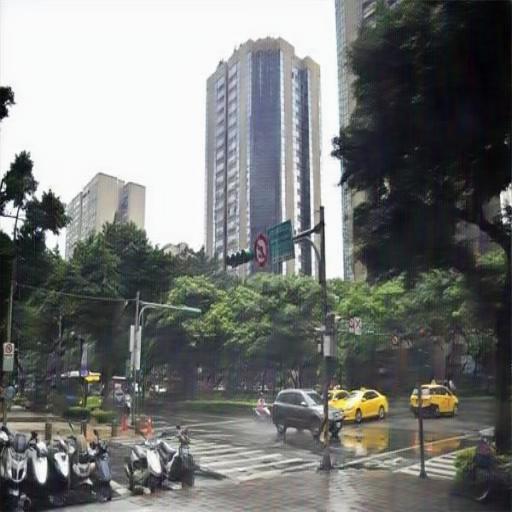}\\
		(a)\hskip 85pt(b)\hskip 85pt(c)\hskip 85pt(d)\hskip 85pt(e)
	\end{center}
	\vskip -18pt \caption{Qualitative results on DDN-SIRR \textbf{real-world} test set. (a) Input rainy image (b)  DID-MDN \cite{Authors18} (c) DDN \cite{Authors17f} (d) SIRR \cite{wei2019semi} (e) Ours.}
	\label{fig:ddnsirr_real}
\end{figure*}

\noindent\textbf{Results on synthetic test set:} The evaluation results on the synthetic test set are shown in Table. \ref{tab:ddnsirr_synthetic}. Similar to \cite{wei2019semi}, we use PSNR as the evaluation metric. We compare the proposed method with several  existing approaches such as DSC \cite{luo2015removing}, LP \cite{li2016rain}, JORDER \cite{yang2017deep}, DDN \cite{Authors17f}, JBO \cite{Authors17c} and DID-MDN \cite{Authors18}. These methods can use only synthetic dataset.  Since the proposed method has the ability to leverage unlabeled real-world data, it is able to achieve significantly better results as compared to the existing approaches. 

Furthermore, we also compare the performance of our method with a recent GMM-based semi-supervised deraining method (SIRR) \cite{wei2019semi}. It can be observed from Table \ref{tab:ddnsirr_synthetic} that the proposed method outperforms SIRR with significant margins. Additionally, we also illustrate the gains\footnote{The gain is computed by subtracting the performance obtained using only $\mathcal{D_L} $ from the performance obtained using $\mathcal{D_L+D_U} $.} achieved due to the use of additional unlabeled real-world data by both the methods. The proposed method achieves greater gains as compared to SIRR, which indicates that the proposed method has better capacity to leverage unlabeled data. 

Qualitative results on the test set are shown in Fig. \ref{fig:ddnsirr_synthetic}. As can be seen from this figure, the proposed method achieves better quality reconstructions as compared to the existing methods.

\noindent\textbf{Results on real-world test set: } Similar to \cite{wei2019semi}, we evaluate the proposed method on the real-world test set from DDN-SIRR. We use no-reference quality metrics NIQE \cite{mittal2012making} and BRISQUE~\cite{mittal2012no} to perform  quantitative  comparison. The results are shown in Table.  \ref{tab:ddnsirr_real}. We compare the performance of our method with SIRR \cite{wei2019semi} which also leverages unlabeled data. It can be observed that the proposed method achieves better performance than SIRR. Note that lower scores indicate better performance. Furthermore, the proposed method is able to achieve better gains with the use of unlabeled data as compared to SIRR.

From these experiments, we can conclude that the proposed GP-based framework when leverages unlabeled real-world data results in better generalization as compared to not using the unlabeled data.

 \begin{table}[t!]
 	\caption{Effect of using unlabeled real-world data in training process on the DDN-SIRR dataset. Evaluation is performed on the \textbf{real-world} test set of DDN-SIRR dataset using no-reference quality metrics (NIQE and BRISQUE). Note that lower scores indicate better performance.}
 	\label{tab:ddnsirr_real}
 	\centering
 	\huge
 	\vskip-10pt
	\resizebox{1\linewidth}{!}{
 	\begin{tabular}{|l|c|ccc|ccc|}
 		\hline
 		\multirow{2}{*}{Metrics} & \multirow{2}{*}{\begin{tabular}[c]{@{}c@{}}Input\end{tabular}} & \multicolumn{3}{c|}{SIRR \cite{wei2019semi}} & \multicolumn{3}{c|}{Ours} \\ \cline{3-8} 
 		&                                                                               & $\mathcal{D_L}$    & $\mathcal{D_L+D_U}$  & Gain  & $\mathcal{D_L}$    & $\mathcal{D_L+D_U}$  & Gain  \\ \hline
 		NIQE                                   & 4.671                                                                         & 3.86   & 3.84     & 0.02  & 3.85   & 3.78     & 0.07  \\
 		BRISQUE                                & 31.37                                                                         & 26.61  & 25.29    & 1.32  & 25.77  & 22.95    & 2.82  \\ \hline
 	\end{tabular}
 } 
\end{table}

 \subsection{Ablation study: SSL experiments}
In this set of experiments, we analyze the capacity of the proposed method to leverage unlabeled data by varying the amount of labeled data used for training the network. Since, the goal is to evaluate the method quantitatively, we use synthetic datasets (Rain800 and Rain200H) for these experiments. Specifically, we run 5 experiments where we train the network on 10\%, 20\%, 40\%, 60\% and 100\% of the dataset as the labeled data $\mathcal{D_L}$. The rest of the dataset is leveraged as the unlabeled data $\mathcal{D_U}$. We use PSNR and SSIM metrics for this ablation study.

The  results on the Rain800 test and Rain200H set are shown in Table \ref{tab:rain800} and \ref{tab:rain100h}, respectively. From these tables, we make following observations: (i) Reducing the amount of labeled data leads to  significant drop in performance as compared to using 100\% of the data as the labeled data. For example, the performance drops from 23.74 dB when using 100\% data to 22.6dB after reducing the labeled data to 40\%. (ii) By using unlabeled data in the proposed SSL framework, we are able to achieve improvements as compared to using only labeled data. (iv) The gain in performance obtained due to the use of unlabeled data is consistent across different amounts of labeled data. (iii) Finally, the proposed method with just 60\% labeled data  (and unlabeled data) is able to achieve performance that is comparable to that achieved by using 100\% labeled data.

 \begin{table}[t!]
 	\centering
 	\caption{SSL experiments on Rain800 \cite{Authors17e} dataset: The percentage of labeled data used for training is varied between 10\% and 100\%. Consistent gains are observed when unlabeled data is leveraged using the proposed method as compared to the use of only labeled data.}
 	\label{tab:rain800}
 	\vskip-9pt
 	\resizebox{1\linewidth}{!}{
 	\begin{tabular}{|l|ccc|ccc|}
 		\hline
 		\multirow{2}{*}{$\mathcal{D_L}$ \%} & \multicolumn{3}{c|}{PSNR}                     & \multicolumn{3}{c|}{SSIM}                      \\ \cline{2-7} 
 		& $\mathcal{D_L}$ & $\mathcal{D_L +D_U}$ & \underline{Gain} & $\mathcal{D_L}$ & $\mathcal{D_L +D_U}$ & \underline{Gain}  \\ \hline
 		10\%                                        & 21.31           & 22.02                & \underline{0.71} & 0.729           & 0.750                & \underline{0.021} \\ 
 		20\%                                        & 22.28           & 22.95                & \underline{0.67} & 0.752           & 0.768                & \underline{0.016} \\ 
 		40\%                                        & 22.61           & 23.60                & \underline{0.99} & 0.761           & 0.788                & \underline{0.027} \\
 		60\%                                        & 22.96           & 23.70                & \underline{0.74} & 0.775           & 0.795                & \underline{0.020} \\ 
 		100\%                                       & 23.74           & --                   & --   & 0.799           & --                   & --    \\ \hline
 	\end{tabular}
}
 \end{table}

\begin{table}[t!]
		\caption{SSL experiments on Rain200H \cite{yang2017deep} dataset: The percentage of labeled data used for training is varied between 10\% and 100\%. Consistent gains are observed when unlabeled data is leveraged using the proposed method as compared to the use of only labeled data.}
		\label{tab:rain100h}
		\vskip-10pt
	\centering
	\resizebox{1\linewidth}{!}{
		\begin{tabular}{|l|ccc|ccc|}
			\hline
			\multirow{2}{*}{$\mathcal{D_L}$ \%} & \multicolumn{3}{c|}{PSNR}                     & \multicolumn{3}{c|}{SSIM}                      \\ \cline{2-7} 
			& $\mathcal{D_L}$ & $\mathcal{D_L +D_U}$ & \underline{Gain} & $\mathcal{D_L}$ & $\mathcal{D_L +D_U}$ &  \underline{Gain}  \\ \hline
			10\%                                        & 22.92           & 23.64                &  \underline{0.72} & 0.742           & 0.767                &  \underline{0.025} \\ 
			20\%                                        & 23.22           & 24.00                &  \underline{0.78} & 0.755           & 0.776                &  \underline{0.021} \\ 
			40\%                                        & 23.84           & 24.75                &  \underline{0.91} & 0.772           & 0.794                &  \underline{0.022} \\ 
			60\%                                        & 24.32           & 25.26                &  \underline{0.94} & 0.782           & 0.808                &  \underline{0.026} \\ 
			100\%                                       & 25.27           & --                   & --   & 0.810           & --                   & --    \\ \hline
		\end{tabular}
	}
	\vskip-10pt
\end{table}

\begin{figure}[t!]
	\begin{center}
		\includegraphics[width=.323\linewidth]{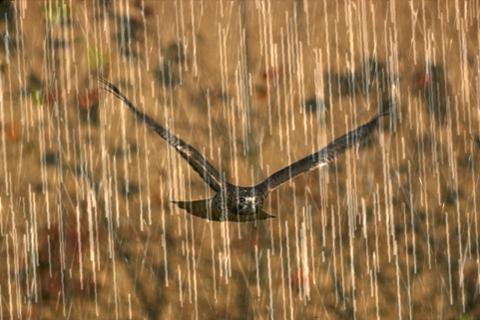}
		\includegraphics[width=.323\linewidth]{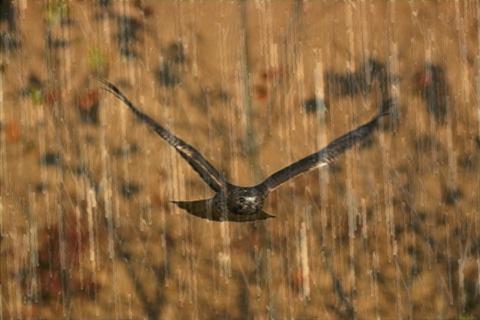}
		\includegraphics[width=.323\linewidth]{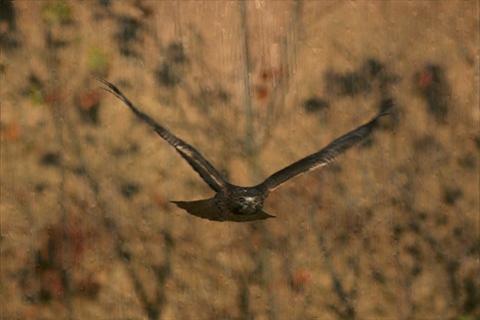}\\
		\includegraphics[width=.323\linewidth]{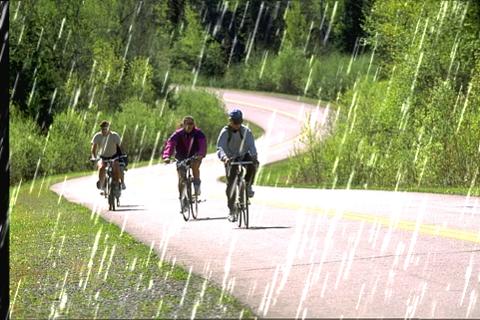}
		\includegraphics[width=.323\linewidth]{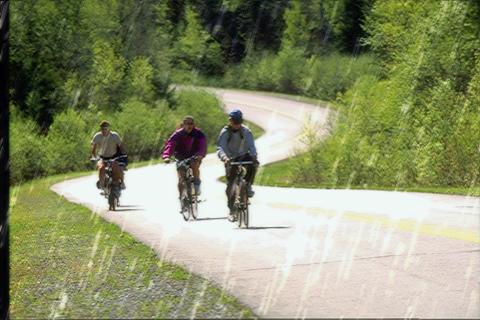}
		\includegraphics[width=.323\linewidth]{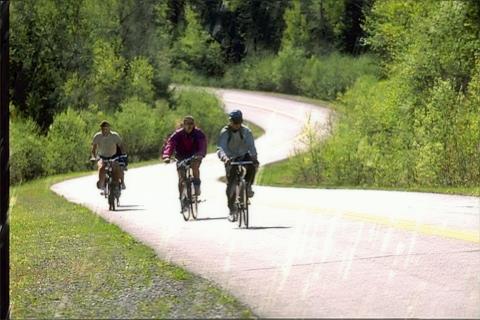}\\
		(a)\hskip 65pt(b)\hskip 65pt(c)
	\end{center}
	\vskip -18pt \caption{Results of experiments with 10\% labeled data on Rain200H (a) Input rainy image (a)  Using only labeled data (c) Using labeled and unlabeled data.}
	\label{fig:rain100h_10}
\end{figure}

\begin{figure}[t!]
	\begin{center}
		\includegraphics[width=.323\linewidth]{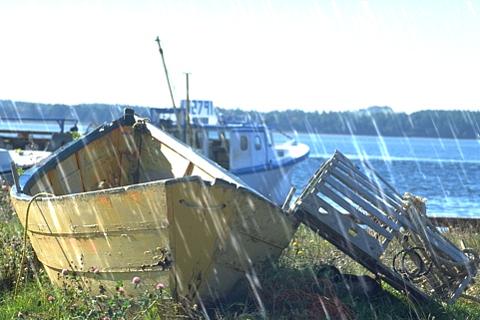}
		\includegraphics[width=.323\linewidth]{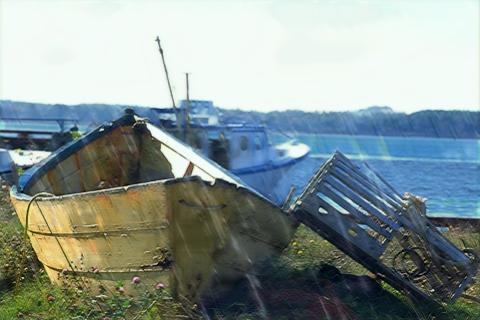}
		\includegraphics[width=.323\linewidth]{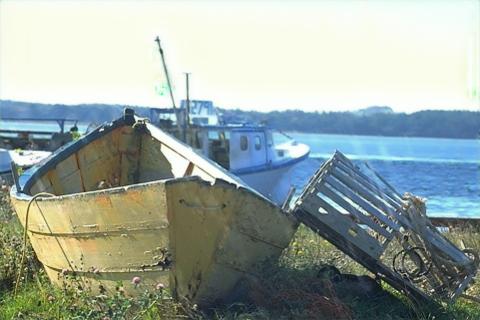}\\
		\includegraphics[width=.323\linewidth]{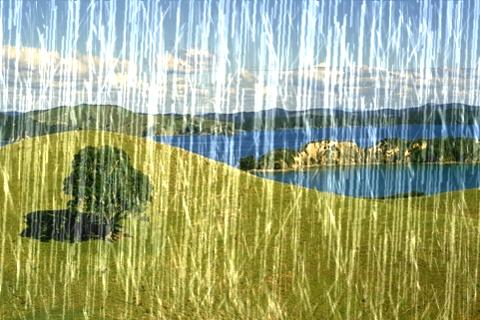}
		\includegraphics[width=.323\linewidth]{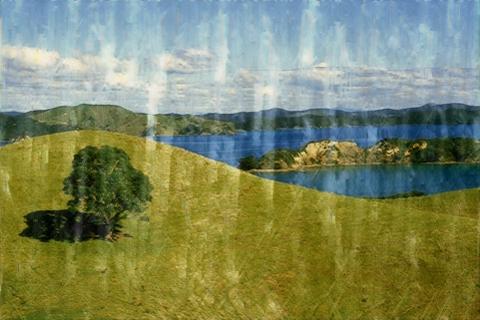}
		\includegraphics[width=.323\linewidth]{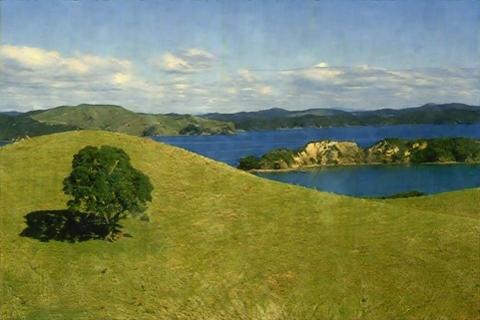}\\
		(a)\hskip 65pt(b)\hskip 65pt(c)
	\end{center}
	\vskip -18pt 
	\caption{Results of experiments with 40\% labeled data on Rain200H (a) Input rainy image (a)  Using only labeled data (c) Using labeled and unlabeled data.}
	\label{fig:rain100h_40}
\end{figure}

Fig. \ref{fig:rain100h_10} and \ref{fig:rain100h_40} show  sample qualitative results when using 10\% and 40\% labeled data, respectively.  It can be observed that  using additional unlabeled data results in better performance as compared to using only labeled data. 

 From these experiments, we can conclude that the proposed method can effectively leverage unlabeled data even with minimal amount of labeled training data. Additionally, only a fraction of the labeled data is sufficient to obtain performance similar to that when using 100\% labeled data.

\subsection{Ablation study: Hyperparameters} 
We also conduct a detailed ablation study to analyze the effects of different hyperparameters present in the proposed method. Due to space constraints, these results along with more qualitative visualizations are provided in supplementary material.

 \section{Conclusion}
We presented a GP-based SSL framework to  leverage unlabeled data during training for the image deraining task. We use supervised loss functions such as $l_1$ and the perceptual loss to train on the labeled data. For unlabeled data, we estimate the pseudo-GT at the latent space  by jointly modeling the labeled and unlabeled latent space vectors using the GP. The pseudo-GT is then used to supervise for the unlabeled samples. Through extensive experiments on several datasets such as Rain800, Rain200H and DDN-SIRR, we demonstrate that the proposed method is able to achieve better generalization by leveraging unlabeled data. 
 

{\small
\bibliographystyle{ieee_fullname}
\bibliography{egbib}
}

\end{document}